\documentclass{article}

\usepackage{arxiv}

\usepackage[english]{babel}

\usepackage[utf8]{inputenc} 
\usepackage[T1]{fontenc}    
\usepackage{amsmath}
\usepackage{graphicx}
  \DeclareGraphicsExtensions{.pdf,.png}
\usepackage{pgfplots}
\pgfplotsset{compat=1.18}

\usepackage{authblk}
\usepackage[colorlinks=true, allcolors=blue]{hyperref}
\usepackage{apacite}
\usepackage[most]{tcolorbox}
\usetikzlibrary{positioning, shapes}
\def\BibTeX{{\rm B\kern-.05em{\sc i\kern-.025em b}\kern-.08em
    T\kern-.1667em\lower.7ex\hbox{E}\kern-.125emX}}
\newcommand{\chatbox}[1]{%
  \begin{tcolorbox}[colframe=black!70!white,colback=white,arc=4pt,boxrule=0.5pt,fontupper=
  \small,top=2pt,left=2pt,right=2pt,bottom=2pt]
  #1
  \end{tcolorbox}
}
\newcommand{\examplebox}[2]{%
  \begin{tcolorbox}[colframe=black!70!white,colback=blue!2!white,arc=4pt,boxrule=0.5pt,fontupper=
  \small,top=2pt,left=2pt,right=2pt,bottom=2pt,title=#1]
  #2
  \end{tcolorbox}
}

\makeatletter
\renewcommand\paragraph{\@startsection{paragraph}{4}{\z@}%
  {3.25ex \@plus1ex \@minus.2ex}%
  {-1em}%
  {\normalfont\normalsize\itshape}}
\makeatother

\begin{document}
\title{An overview of model uncertainty and variability in LLM-based sentiment analysis. Challenges, mitigation strategies and the role of explainability}

\newcommand{\shorttitle}{Overview of model uncertainty and variability in LLM-based sentiment analysis} 

\author{David Herrera-Poyatos$^{1,*}$, Carlos Pel\'aez-Gonz\'alez$^1$, Cristina Zuheros$^1$, Andr\'es Herrera-Poyatos$^1$, Virilo Tejedor$^1$, Francisco Herrera$^1$, Rosana Montes$^2$
}

\affil{$^1$Department of Computer Science and Artificial Intelligence, Andalusian Institute of Data Science and Computational Intelligence (DaSCI), University of Granada, Spain. \\ Emails: \texttt{\{divadhp, carlosprog, czuheros, andreshp\}@ugr.es}, \texttt{virilo@gmail.com},  \texttt{herrera@decsai.ugr.es} \\
$^{*}$Corresponding author. }

\affil{$^2$Department of Software Engineering, Andalusian Institute of Data Science and Computational Intelligence (DaSCI), University of Granada, Spain. \\ Email: \texttt{rosana@ugr.es}}

\date{\today}

\maketitle
\begin{abstract}
Large Language Models (LLMs) have significantly advanced sentiment analysis, yet their inherent uncertainty and variability pose critical challenges to achieving reliable and consistent outcomes. This paper systematically explores the Model Variability Problem (MVP) in LLM-based sentiment analysis, characterized by inconsistent sentiment classification, polarization, and uncertainty arising from stochastic inference mechanisms, prompt sensitivity, and biases in training data. We analyze the core causes of MVP, presenting illustrative examples and a case study to highlight its impact. In addition, we investigate key challenges and mitigation strategies, paying particular attention to the role of temperature as a driver of output randomness and emphasizing the crucial role of explainability in improving transparency and user trust. By providing a structured perspective on stability, reproducibility, and trustworthiness, this study helps develop more reliable, explainable, and robust sentiment analysis models, facilitating their deployment in high-stakes domains such as finance, healthcare, and policymaking, among others.
\end{abstract}


\section{Introduction}
Sentiment analysis has emerged as a critical application of large language models (LLM) in fields such as customer feedback analysis, financial market predictions, brand reputation monitoring, and trend detection on social media. Traditional sentiment analysis approaches relied on rule-based lexicons or supervised machine learning models, which, while interpretable, struggled with nuanced expressions such as sarcasm, irony, and contextual sentiment shifts \cite{wankhade2022survey,zhang2023sentimentanalysiseralarge, krugmann2024-SAintheageofGenAI}. With the introduction of LLMs such as GPT-4, BERT, RoBERTa, and T5, sentiment classification has improved significantly in terms of precision, contextual understanding, and adaptability to various domains. LLMs leverage their vast pre-training corpora and deep transformer architectures to understand sentiment beyond simple polarity detection, incorporating emotion classification, aspect-based sentiment analysis, and entity-level sentiment extraction \cite{yang2024large}.

Despite these advancements, the reliance on probabilistic text generation and deep feature representations introduces challenges related to output variability, inconsistency between inference runs, and susceptibility to biases in training data \cite{beigi2024rethinking}. Unlike traditional classifiers, which yield deterministic output, LLMs can generate different sentiment scores for the same input based on factors such as decoding parameters, prompt phrasing, and the model's internal confidence in its predictions. This variability is particularly concerning in high-stakes decision-making applications, such as automated financial sentiment analysis, where unstable predictions can lead to inaccurate market forecasts. Addressing this issue requires robust techniques such as uncertainty quantification, model calibration, and ensemble averaging to enhance stability, reliability, and explainability in sentiment classification.

The Model Variability Problem (MVP) refers to the phenomenon in which an LLM or machine learning system produces inconsistent outputs for the same input in multiple runs \cite{wankhade2022survey}. This issue arises in various natural language processing applications but is particularly problematic in sentiment analysis, where a model tasked with assigning a sentiment polarity score (ranging from 0 to 1) may yield different values for identical input text. These inconsistencies result from the stochastic nature of LLM inference mechanisms, leading to fluctuations that impact the reliability, trustworthiness, and downstream applications of the model in decision-making systems.

The survey \cite{wankhade2022survey} published in 2022, entitled \textit{"A Survey on Sentiment Analysis Methods, Applications, and Challenges"}, pay attention to the uncertainty and the MVP. It aligns closely with the uncertainty and variability described in the context of LLM-based sentiment analysis. The authors discuss key challenges such as domain dependency, ambiguity in textual data, implicit language understanding (including sarcasm and irony), and feature selection complexities ---all contributing factors to variability issues observed in modern sentiment analysis approaches. These identified challenges echo the broader MVP, highlighting fundamental issues like aleatoric uncertainty due to ambiguous language and epistemic uncertainty caused by insufficient domain knowledge or lack of generalization capabilities in models. In addition, the survey emphasizes the limitations of conventional sentiment analysis methods, including lexicon-based and supervised approaches, stressing that each method faces difficulties in reliably capturing nuanced sentiment, particularly in real-world settings involving sarcasm, irony, slang, and domain-specific terminology. This is directly related to MVP, as similar ambiguities significantly affect LLM predictions, causing inconsistent sentiment classifications between different inference runs \cite{da2025understanding}.

Recent advances in LLMs have significantly impacted sentiment analysis, notably enhancing sentiment analysis-based crowd decision making (CDM) by using structured prompt design strategies, as evidenced by recent empirical studies utilizing different LLMs \cite{Herrera-Poyatos2025large}. Although these methods show promising potential for extracting consensus-driven sentiment classifications from large-scale opinion datasets, the inherent uncertainty and variability within LLMs pose fundamental challenges. The variability problem, which arises from sensitivity to prompt variations, stochastic inference methods, and biases of training datasets, critically impacts the reliability and consistency of crowd-driven sentiment classification. This paper extends the existing analysis by exploring the uncertainty and variability factors that affect the accuracy of sentiment classification, highlighting the need for refined methodologies that address consistency, robustness, and transparency in CDM contexts supported by LLM.

In addition, given the high stakes involved in CDM and other critical applications of sentiment analysis, particular emphasis must be placed on the explainability and interpretability of LLM-generated sentiment predictions and explanations. As discussed extensively in the recent literature, the opaque nature of LLMs impedes understanding of model predictions, posing significant barriers to trust and user acceptance. Herrera in \cite{herrera2025} underscores the importance of adopting explainability frameworks to improve the transparency of AI-driven sentiment analysis, facilitating better human-AI interaction, trust building, and informed decision-making. Thus, this paper also explores strategies to integrate robust explainability methodologies, ensuring that sentiment classification outputs are not only consistent and reliable but also transparent and comprehensible for diverse stakeholders involved in CDM contexts.

The consequences of model variability in sentiment analysis and other nature language processing applications include:

\begin{itemize}

   \item Unstable sentiment classification: A business analyzing customer feedback may receive conflicting sentiment scores from the same input.

   \item Bias amplification: Variability can exacerbate inherent model biases, leading to systematic errors in human-AI decision making, where AI systems must assist the human with advice. 

   \item Reduced reproducibility in sentiment analysis with LLMs: Studies relying on LLM output may not reproduce results, affecting model benchmarking.

   \item Challenges in trustworthy AI: End users and policymakers require an explainable and consistent AI behavior, which variability undermines.
  
\end{itemize}

The purpose of this paper is to provide a comprehensive and holistic analysis of MVP and associated uncertainties that arise from the use of LLM-based sentiment analysis. We pay attention to challenges, mitigation strategies, and the role of explainability.

We adopt a structured approach, beginning with some illustrative examples to demonstrate variability in LLM-generated sentiment predictions. Subsequently, we performed an in-depth analysis of the primary factors contributing to uncertainty and variability, focusing on critical aspects such as stochastic inference methods, prompt sensitivity, and biases in training data. Finally, the article identifies emerging trends and key challenges and describes promising directions and methodologies as mitigation strategies. They include uncertainty quantification, ensemble consensus approaches, and robust explainability frameworks, to improve reliability, stability, and transparency in LLM-based sentiment analysis applications, among others.

In order to do that, the paper is organized as follows. Section \ref{sec:Case} shows some illustrative examples on the problem of focus, uncertainty, and variability. Section \ref{sec:Ten} introduces the fundamental reasons for model variability with a literature analysis. Section
\ref{sec:Exp} introduces a reflection and analysis on the importance of explainability for LLM. Section \ref{sec:Challenges} discusses challenges and mitigation strategies for MVP in LLM-based sentiment analysis. Finally, some conclusions are pointed out in Section \ref{sec:Con}. 

\section{Illustrative examples to show the uncertainty and model variability in LLMs-based sentiment analysis}
\label{sec:Case}

It is crucial to present illustrative examples when addressing uncertainty and MVP in LLM-based sentiment analysis, as tangible demonstrations facilitate clearer visualization and a deeper understanding of this complex phenomenon. Through concrete scenarios, stakeholders can directly observe how variability arises, the degree to which it affects model predictions, and why it constitutes a significant challenge in practical applications. Real-world examples offer an accessible entry point for grasping abstract concepts such as stochastic inference, prompt sensitivity, or subtle contextual shifts, enabling more effective discussions, diagnostics, and ultimately the formulation of robust mitigation strategies.

To address these challenges, we present two case studies in Subsection \ref{sec:dataset} that vividly illustrate the unpredictable nature of model outputs when dealing with real-world sentiment data. In the first case of study, we explore variability in repeated sentiment evaluations using the GPT-4o model, demonstrating how the same input can yield fluctuating predictions due to stochastic inference. In the second case of study, we investigate inconsistencies between numerical sentiment scores and categorical labels obtained from the Mixtral 8x22B model, revealing the impact of prompt sensitivity and contextual interpretation on model reliability.

Following these case studies, we discuss the broader implications of model variability through additional illustrative examples in Section \ref{sec:examples}. These examples further emphasize how unpredictable and inconsistent sentiment predictions can undermine the credibility of LLM-based applications, particularly in high-stakes contexts such as finance, healthcare, and customer experience management. By presenting both empirical evidence and conceptual scenarios, our aim is to foster a deeper understanding of the critical challenges posed by uncertainty in sentiment analysis and advocate robust strategies to mitigate model variability and enhance trustworthiness.

\subsection{Case of study: TripR-2020Large dataset}\label{sec:dataset}

In order to introduce a short analysis, we consider the TripR-2020Large dataset~\cite{zuheros2021sentiment} since it collects real data to evaluate Decision-Making (DM) models with unrestricted natural language input. The TripR-2020Large dataset\footnote{\url{https://github.com/ari-dasci/OD-TripR-2020Large}} contains 474 written reviews in English from 132 TripAdvisor users, who express their experiences in four restaurants, forming the set of alternatives: $X = \{x_{1}, x_{2}, x_{3}, x_{4}\}$ = $\{${\textit{The Oxo Tower}}, {\textit{The Wolseley}}, {\textit{The Ivy}}, {\textit{J. Sheekey}}$\}$. Since not all experts review every restaurant, the dataset comprises slightly fewer than $132\times4$ documents.

In the following, we present an analysis of the opinion introduced in Figure \ref{fig:review-376} and show the uncertainty observed when the query is repeated 100 times using the GPT 4o model. This analysis highlights how variability manifests itself in model predictions and quantifies the inherent uncertainty associated with sentiment analysis. To conduct this study, we used the following prompt to assess the sentiment of the given review:

\chatbox{Rate the sentiment of this review on a continuous scale from 0 to 1, where 0 means entirely negative, and 1 means entirely positive. The answer must be only a number:\newline [DOCUMENT]}

\begin{figure}[h!]
    \centering
    \begin{tikzpicture}
        \begin{axis}[
            ybar,
            bar width=1.0cm,
            width=0.8\linewidth,
            height=0.5\linewidth,
            xlabel={Score},
            ylabel={Frequency},
            title={Frequency of Scores},
            ymin=0,
            xtick=data,
            nodes near coords,
            enlarge x limits=0.1,
            xticklabel style={rotate=45, anchor=east},
            every axis plot/.append style={fill=cyan, opacity=0.7}
        ]
        \addplot coordinates {(0.3, 3) (0.35, 3) (0.4, 63) (0.45, 8) (0.5, 18) (0.55, 1) (0.6, 4)};
        \end{axis}
    \end{tikzpicture}
    \caption{Uncertainty under the opinion analysis using GPT 4o.}
    \label{fig:uncertainty}
\end{figure}
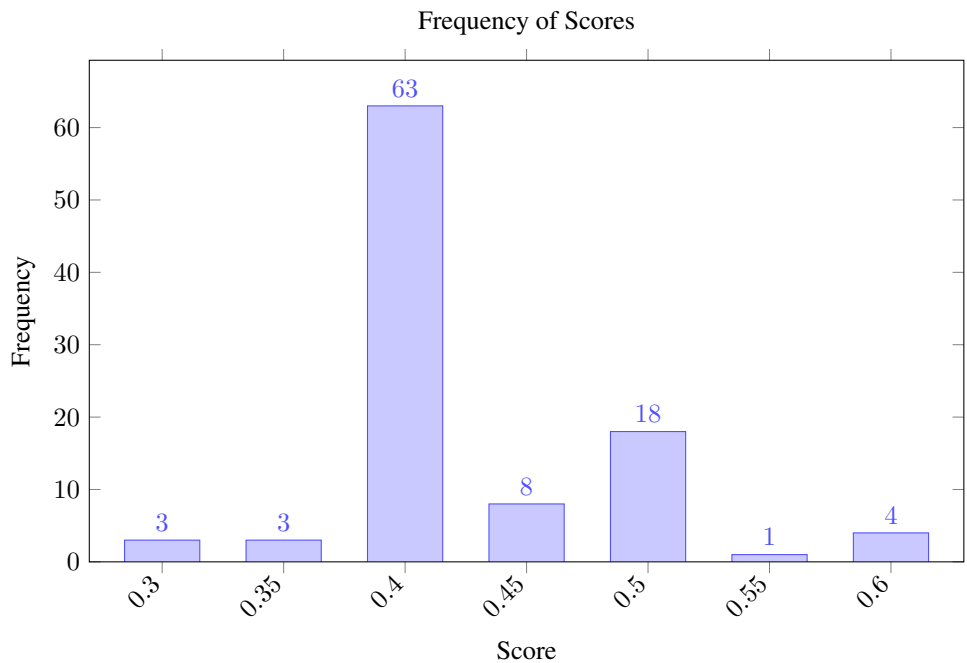

The resulting variability distribution, shown in Figure \ref{fig:uncertainty}, shows how the predictions fluctuate even when analyzing the same review multiple times. It should be noted that it fluctuates between negative (0.3) and positive (0.6) values. 

\begin{figure}[h!]
    \centering
    \examplebox{Sample review}{The Ivy almost came up to all our expectations and the 'hype'. Excellent 'front of house', tables far enough apart so that 'next door's' elbows not clashing with ours; first class waiter service (although we had to ask for our pre meal drinks twice from the wine waiter!); food exceptional with Brixham Scallops and Claves Livers outstanding. However, minimum price of wine @ £32 somewhat excessive for a label that could be bought in supermarket for £9 and 'The NOISE!!!!!!!!!!!!!!!!!!!!!!!!!!!!!!!!!!!!!!'Even at 10.30 the place was packed and it sounded like a good night (although very rare no doubt) when England were winning at Wembley. Reasonable conversation was impossible. Everyone seemed to be shouting at each other to be heard which was virtually impossible because of the adjacent table of Americans who seemed not to realise that the people they were talking to (or rather shouting at) were sitting next to them. One of the worst examples of 'I'm payin' buddy so I can dominate the whole restaurant 'cos my wallet's bigger than yours and I can laugh louder than you'No extent of 'Maggie Smith' looks or good old British 'tutting' seemed to get through to these boorish, thick skinned morons. The staff, even though they could see how annoyed people were' seemed to timid or frightened to do anything.When they got up and left a corporate sigh of relief was audible.Perhaps the management could install a decibel meter and ask for quiet when 125 Db is reached}
    \caption{Review from the TripR-2020Large dataset.}
    \label{fig:review-376}
\end{figure}

\begin{figure}[h!]
    \centering
    
    \includegraphics[width=0.80\textwidth]{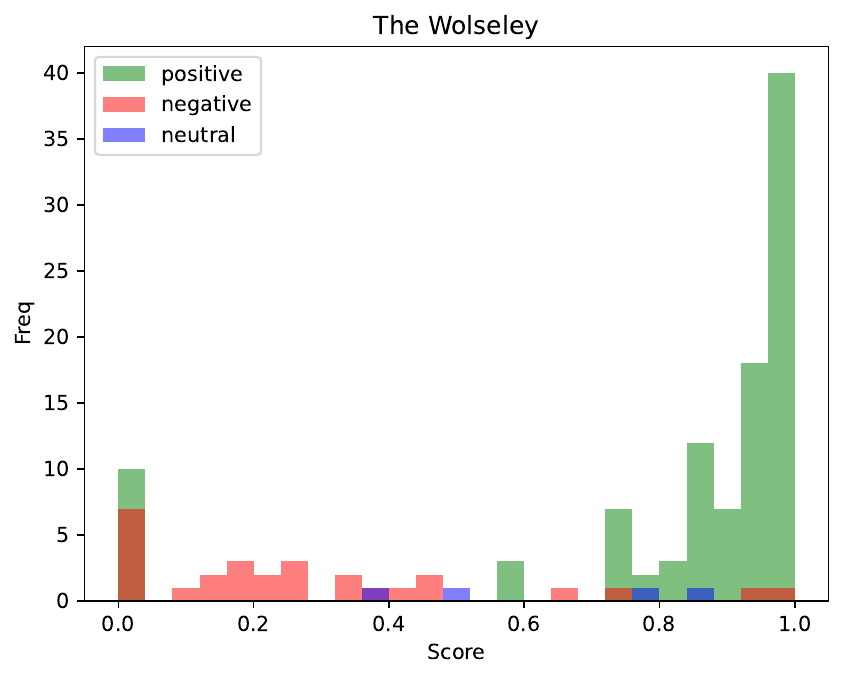}
    \caption{Inconsistency between numerical and linguistic polarity using Mixtral 8x22B model}
    \label{fig:inconsistency}
\end{figure}

The second study, shown in   Figure \ref{fig:inconsistency}, presents the evaluation of all opinions for the restaurant \textit{The Wolseley},  obtained using the Mixtral 8x22B model. This study aims to analyze the consistency between numerical sentiment scores and categorical labels, highlighting the inherent challenges posed by prompt sensitivity and contextual variability.

The data shown in Figure \ref{fig:inconsistency} consists of the following elements:

\begin{itemize}
    \item The histogram illustrates the frequency of sentiment scores ranging from 0.0 (entirely negative) to 1.0 (entirely positive), with intermediate values indicating varying degrees of sentiment polarity. 
    \item The color of each bar represents the sentiment label predicted by the same model, obtained by using a different prompt that explicitly asks for a label among \textit{positive}, \textit{neutral} or \textit{negative}. Therefore, the color coding reflects the prediction of the sentiment label, while the height of each bar corresponds to the frequency of scores.
\end{itemize}

The histogram is generated by combining the results from two different prompts, which are as follows:

\chatbox{Classify the sentiment of the following text as positive, neutral or negative, the answer must be a single label and one word:\newline [DOCUMENT]}

 \chatbox{Classify the sentiment of the following text using a score between 0 and 1, where 0 represents a completely negative sentiment and 1 represents a completely positive sentiment. The answer must be only a number:\newline [DOCUMENT]}

The responses obtained from these prompts are then matched for each review, allowing both the model's score and label predictions to be visualized together in the histogram.

 Figure \ref{fig:inconsistency} reveals significant inconsistencies between the numerical sentiment score and the categorical label. For example, some review classified as negative exhibit relatively high sentiment scores, while positive labels occasionally appear at low scores. This discrepancy highlights the model's variability, reflecting challenges related to stochastic inference and prompt sensitivity.

The two presented case studies clearly demonstrate the profound challenges posed by uncertainty and model variability in LLM-based sentiment analysis. The first study highlights the inherent instability of model outputs when repeating the same sentiment analysis query multiple times, revealing how even minor stochastic variations can lead to significantly different predictions. This phenomenon is particularly concerning when consistent input should logically yield stable output, but the inherent randomness of the model results in a fluctuating range of sentiment scores.

The second study exposes an equally critical issue: the inconsistency between numerical sentiment scores and categorical labels. Even when using the same model and evaluating the same input with slightly different prompts, the outputs diverge significantly, revealing a lack of coherence between quantitative and qualitative sentiment assessment. This inconsistency points to a fundamental challenge in how LLMs interpret and classify sentiment, especially when prompt phrasing subtly alters the context or interpretation.

Together, these case studies reveal that model variability and inconsistency are not merely occasional glitches but rather systematic challenges that arise from the nature of LLM-based sentiment analysis. Such variability significantly undermines the reliability and trustworthiness of automated sentiment classification, particularly in critical applications such as customer feedback analysis, healthcare monitoring, and financial sentiment prediction. Addressing these challenges demands not only a deeper understanding of how variability emerges but also the development of robust strategies to enhance model consistency and reliability.

By quantifying the extent of variability and demonstrating its impact through real-world examples, we highlight the urgent need for robust mitigation techniques that can reduce the unpredictability of model outputs and enhance the stability of sentiment predictions in practical applications.

\subsection{Illustrative examples}\label{sec:examples}
Illustrative examples can vividly showcase the practical consequences of model variability, particularly in high-stakes domains such as finance, healthcare, and consumer analytics, among others. When variability is demonstrated through clear cases, such as analyzing sentiment from customer reviews or interpreting financial news headlines, users and developers alike can better appreciate its impact on reliability and trustworthiness. This practice not only helps to recognize the urgent need for stable and transparent AI solutions but also emphasizes the importance of investing in research to develop techniques for reducing MVP, thus enhancing the overall dependability and effectiveness of sentiment-analysis applications powered by LLMs.

\paragraph{Sentiment variability in customer reviews}
To clearly illustrate the phenomenon of model variability in sentiment analysis, consider an example involving customer reviews analyzed by GPT-4. When analyzing the sentence, \textit{"Oh great, another rainy day!"}, GPT-4 may interpret this sentiment differently across multiple inference runs. In one instance, influenced by the literal meaning of \textit{"great"}, it might assign a neutral or slightly positive sentiment score, whereas in another instance, detecting potential sarcasm, it could produce a strongly negative sentiment. Such inconsistencies reflect the inherent uncertainty that stems from the ambiguity of natural language and the stochastic decoding processes employed by LLMs, significantly affecting the reliability of sentiment classifications in practical, real-world applications.

\paragraph{Variability in financial sentiment analysis} 
Another illustrative example is observed in financial sentiment analysis tasks using LLMs such as ChatGPT, where predictions about market sentiments based on news headlines exhibit considerable variability. For instance, the headline \textit{"Company X announces a surprise merger"} might trigger diverse sentiment polarity scores across different runs, varying from cautiously optimistic to highly positive. This variation arises due to differences in prompt construction, subtle contextual interpretations, and model parameter settings such as temperature or top-k sampling, leading to instability that critically impacts decision-making in scenarios requiring precision and reproducibility, such as algorithmic trading or risk assessment. These examples underscore the urgent need to understand and mitigate MVP to improve trust and effectiveness in LLM-based sentiment analysis.

\section{ A dozen fundamental reasons for model variability}
\label{sec:Ten}

 MVP refers to the phenomenon where LLMs produce inconsistent outputs for the same input on multiple runs. Based on the analysis of key literature, we identify a dozen fundamental reasons that contribute to this issue. We provide a short introduction to each and the appropriate literature that supports them.  

\begin{enumerate}

\item \textit{Aleatoric and epistemic uncertainty.}  

Uncertainty in sentiment classification can be categorized into two primary types: aleatoric uncertainty, which arises from inherent randomness in data, and epistemic uncertainty, which stems from knowledge limitations within the model. These factors contribute significantly to the MVP, leading to inconsistent sentiment predictions across inference runs.

Aleatoric uncertainty manifests itself when textual data contains ambiguities, sarcasm, or sentimentally mixed expressions, making the interpretation highly dependent on context. Sentiment analysis models often struggle with these complexities, leading to unstable and inconsistent sentiment classifications. Addressing aleatoric uncertainty requires enhanced contextual embeddings, advanced linguistic modeling techniques, and probabilistic output representations to better handle ambiguous textual inputs. Studies such as \cite{shorinwa2024survey} identify data-driven noise ---stemming from annotation inconsistencies, ambiguous labels, and linguistic variability--- as a primary source of aleatoric uncertainty in LLM-based sentiment analysis. Furthermore, \cite{beigi2024rethinking} highlights how social media slang, domain shifts, and informal text variations exacerbate this uncertainty, making it challenging for models to generalize sentiment classification across different contexts.

Epistemic uncertainty, on the other hand, arises when gaps in pre-training data prevent the model from confidently handling unfamiliar or underrepresented linguistic structures. This type of uncertainty leads to unstable predictions, particularly in domain-specific sentiment tasks where LLMs lack sufficient exposure to nuanced vocabulary. Furthermore, epistemic uncertainty can result in confidence misalignment, where models express unwarranted certainty in incorrect predictions, undermining trustworthiness. \cite{reveilhac2024chatgpt} explores how knowledge limitations in LLM-powered voting systems introduce contradictions in sentiment-based decision-making, reinforcing the need for uncertainty-aware training approaches. Similarly, \cite{passerini2025fostering} shows how human-LLM interactions can mitigate or amplify epistemic uncertainty, depending on whether the model is trained on high-quality, diverse sentiment data or exposed to biased, repetitive user input.

Addressing both aleatoric and epistemic uncertainty requires a multifaceted approach, integrating data augmentation techniques, uncertainty-aware learning frameworks, and structured fine-tuning methodologies. By enriching training data, improving contextual sensitivity, and implementing confidence calibration techniques, LLMs can achieve greater stability and reliability in sentiment classification. Future research must focus on quantifying these uncertainties systematically and designing adaptive models that can dynamically adjust confidence levels based on input complexity.

\item \textit{The role of temperature in LLM output variability.} 

One of the most influential hyperparameters yet examined in LLM is the \textit{temperature}, which directly controls the stochasticity of the output generation process. Temperature scales the logits (output probabilities) before applying the softmax function, thus modulating how deterministic or exploratory the model's sampling behavior becomes during inference. Lower temperatures (e.g., $T=0.1$--$0.3$) make the model output more deterministic by increasing the probability mass on the most likely tokens, while higher temperatures (e.g., $T=0.8$--$1.5$) introduce more randomness, promoting diversity and creativity at the expense of consistency.

This parameter has critical implications for the MVP. High-temperature settings, although useful in open-ended tasks like creative writing or brainstorming, inherently increase output variance—even for semantically equivalent prompts. This introduces unpredictability and reduces reliability in use cases where stability, reproducibility, and fairness are essential, such as sentiment analysis, medical decision support, or legal QA. In such contexts, repeated queries with identical prompts may yield divergent responses, undermine user trust, and compromise decision integrity.

Recent studies have shown that even at moderate temperature settings ($T=0.7$), LLMs such as GPT-3.5, Falcon, or LLaMA exhibit significant variance in sentiment polarity, justification styles, and factuality levels \cite{beigi2024rethinking}. This variance becomes particularly problematic in applications relying on aggregate decision models (e.g., crowd decision-making or sentiment voting systems), where fluctuations in individual model predictions can distort final consensus or rankings. Moreover, the interaction between temperature and prompt sensitivity exacerbates MVP: small syntactic rephrasings can drastically shift the model sampling trajectory under high-temperature decoding.

Understanding and controlling the effects of temperature is thus vital not only for task-specific performance but also for broader goals in LLM trustworthiness, explainability, and reproducibility.

\item \textit{Inference stochasticity in inference and sampling mechanisms}. 

 MVP in LLM-based sentiment analysis is exacerbated by stochastic inference mechanisms that introduce non-deterministic behavior into sentiment predictions. LLMs utilize probabilistic decoding strategies such as temperature scaling (previously analyzed), top k sampling, and beam search to generate responses. Although these methods improve response diversity and adaptability, they also lead to inconsistent sentiment outputs, even when processing identical inputs multiple times. This variability poses significant challenges in applications that demand stability and reproducibility, such as financial sentiment analysis, legal document evaluation, and automated decision-making systems.

One of the main contributors to sentiment variability in LLMs is the randomness of token selection during inference, as evidenced in studies such as \cite{ye2024benchmarking}. The authors analyze how sampling randomness impacts sentiment classifications, demonstrating that identical sentiment analysis tasks can yield inconsistent results due to stochastic decoding. The study specifically examines temperature scaling and top-k sampling, highlighting how these hyperparameters influence the distribution of possible sentiment labels. Similarly, \cite{lefort2024} identifies quantile-based variations in sentiment classification across repeated runs, further confirming that LLM inference introduces an inherent degree of unpredictability into sentiment analysis.

Furthermore, research by \cite{loya2023exploring} explores how hyperparameter sensitivity impacts model decision making, emphasizing that even when inference settings are held constant, minor prompt variations can still lead to differing sentiment classifications. This highlights an essential issue: stochastic variability is not just a function of temperature or top k sampling, but also depends on contextual prompt dependencies, making the problem even more complex. The findings of \cite{atil2024llm} further validate this concern, demonstrating accuracy fluctuations of up to 10\% across repeated identical inference runs, even in cases where deterministic configurations were enforced. Their study introduces stability-focused evaluation metrics, including the Total Agreement Rate at N (TARr@N and TARa@N), which systematically measure inference instability across different sentiment classification tasks.

Beyond stochastic inference, uncertainty quantification techniques play a crucial role in mitigating variability in sentiment classification. As highlighted in \cite{ji2025overview}, epistemic and aleatoric uncertainty significantly impact the trustworthiness of LLM sentiment predictions. The study proposes uncertainty estimation based on entropy, semantic consistency checks, and confidence-aware calibration techniques, which help mitigate inconsistencies by quantifying and adjusting model confidence in uncertain sentiment classifications. These statistical methodologies are especially relevant in high-stakes sentiment-driven applications, where even slight variations in LLM sentiment output could have significant financial, legal, or societal implications.

To counteract the effects of inference stochasticity and enhance stability in sentiment classification, recent studies propose multiple mitigation strategies. \cite{atil2024llm} suggests integrating ensemble-based averaging techniques, which aggregate the outputs of multiple inference runs to reduce variability and reinforce the stability of classification. Similarly, confidence calibration techniques, as explored in \cite{ye2024benchmarking}, provide mechanisms to align model predictions with uncertainty quantification, reducing the probability of sentiment fluctuations due to stochastic effects. These findings reinforce the need for systematic stability-aware evaluation frameworks in sentiment analysis, ensuring that LLM-based models maintain consistency and reliability in real-world applications.

\item \textit{Prompt sensitivity.}  

One of the primary factors contributing to the MVP in LLM-based sentiment analysis is prompt sensitivity, where even minor variations in input phrasing can lead to significant shifts in sentiment classification. This phenomenon underscores the nondeterministic nature of LLMs, which arises from their reliance on stochastic decoding mechanisms such as temperature scaling, top-k sampling, and beam search. Although these techniques improve flexibility and adaptability, they also introduce unpredictability in sentiment predictions, making it challenging to achieve consistent and reproducible results across inference runs. Addressing prompt sensitivity requires careful prompt engineering, structured input standardization, and fine-tuning approaches to reduce variability and ensure more stable sentiment predictions.

Several studies have shown that prompt formulation significantly impacts LLM output variability, leading to inconsistent sentiment classifications in AI-driven applications. The study \cite{yang2024llm} explores how prompt variation affects voting behaviors in LLMs, revealing that subtle changes in question phrasing or persona framing can shift collective AI-generated decisions. This aligns with findings in sentiment analysis, where slight modifications in a prompt can cause a sentiment label to fluctuate between positive, neutral, or negative, highlighting the instability introduced by linguistic framing in LLM-based sentiment classification systems. This sensitivity not only impacts reproducibility, but also raises concerns about the reliability of AI-driven decision-making frameworks in high-stakes applications such as financial analysis, legal evaluations, and policy making.

Moreover, recent studies confirm that LLM decision-making behavior is highly susceptible to prompt variation, further validating the claim that sentiment analysis models exhibit inconsistency depending on input wording \cite{loya2023exploring}. A detailed analysis in \cite{zhang2023sentimentanalysiseralarge} demonstrates that sentiment polarity fluctuates significantly on prompt construction, making model output highly unpredictable. Even small changes in phrasing, word emphasis, or contextual framing can lead to drastic sentiment classification changes, emphasizing the importance of structured prompt formulation. Similarly, \cite{krugmann2024-SAintheageofGenAI} highlights that models such as GPT-3.5 and GPT-4 produce varying sentiment predictions based on prompt specificity, with explicitly structured prompts yielding greater consistency, while vague or ambiguous inputs amplify the variability of sentiment classification.

Beyond sentiment analysis, research on human-LLM interaction modes further illustrates the impact of different prompting strategies on model output. The taxonomy proposed in \cite{gao2024taxonomy} categorizes interaction techniques into standard prompting, UI-enhanced interactions, context-driven inputs, and agent-facilitated prompting, each of which can contribute to variability in LLM sentiment classification. Understanding these structured interaction paradigms is essential to design consistent, robust, and reproducible sentiment analysis methodologies that reduce fluctuations in model output.

In addition, prompt sensitivity has been shown to introduce significant uncertainty in AI-driven recommendation systems. The recent study~\cite{kweon2025uncertainty} highlights that LLM-based recommendation systems face substantial volatility and uncertainty due to prompt sensitivity, variations in user history length, and stochastic inference methods, even when identical input conditions are maintained. To address this, they propose an uncertainty quantification framework that measures reliability in AI-generated recommendations and decomposes uncertainty into two key dimensions:
\begin{itemize}
    \item Recommendation uncertainty – intrinsic ambiguity due to the complexity of the recommendation task itself.
     \item Prompt uncertainty – variability arising specifically from differences in prompt formulations, reinforcing the need for structured input standardization.
\end{itemize}

Prompt sensitivity is particularly critical in LLM-driven collective decision-making tasks, where LLMs act as proxies for individual opinions. Their responses can fluctuate significantly based on prompt structure and contextual framing, leading to inconsistencies in aggregated sentiment classifications. This variability is especially concerning in applications where LLMs aggregate diverse viewpoints to form a consensus, as different prompt designs may yield divergent sentiment scores, ultimately affecting overall model reliability. To minimize LLM sensitivity in sentiment classification tasks, \cite{jarrett2025language} emphasize the need for robust prompt engineering strategies and structured input standardization techniques.

 \item \textit{Domain-specific challenges.} 
 
MVP is further amplified in domain-specific sentiment analysis, where general-purpose LLMs struggle to adapt to specialized fields such as legal, financial, and medical domains. Domain-specific challenges arise from the inherent complexities, constraints, and specialized requirements of a particular field. These models are typically trained on broad and diverse corpora, which may not provide sufficient coverage of the nuanced vocabulary, terminology, and contextual cues specific to certain disciplines. As a result, LLMs frequently misinterpret domain-specific expressions, leading to inconsistent or unstable sentiment predictions when applied to specialized tasks. This variability underscores the need for domain-adapted training, fine-tuning methodologies, and hybrid modeling strategies to improve the accuracy, robustness, and reliability of LLMs in specialized sentiment classification tasks.

A key issue in domain-specific sentiment analysis is domain drift, where LLMs trained on general datasets fail to generalize effectively to specialized applications. As highlighted in \cite{van2025advantages}, LLMs exhibit greater instability when applied to finance, healthcare, and legal analysis, as their probabilistic nature often misinterpret technical language and domain-specific sentiment cues. The study argues that lexicon-based sentiment models, which rely on predefined sentiment rules, provide greater stability in structured environments where deterministic rules better capture sentiment classification nuances.

Similarly, \cite{zhang2023sentimentanalysiseralarge} finds that LLMs trained in mixed-domain datasets exhibit higher variability when performing sentiment classification in specialized fields. The study reveals that models trained without domain-specific adaptation struggle to interpret context-dependent terminology, leading to fluctuations in sentiment classification even when presented with semantically similar inputs. This suggests that cross-domain generalization remains a persistent challenge, requiring adaptive fine-tuning, domain-specific data augmentation, and lexicon-enhanced hybrid models to stabilize LLM-based sentiment analysis.

One promising approach to mitigating domain-specific variability is the integration of hybrid models that combine lexicon-based approaches with LLM-driven sentiment classification. Using the interpretability of predefined sentiment rules with the contextual flexibility of LLMs, hybrid frameworks can achieve greater consistency and robustness across domain-specific sentiment tasks. Furthermore, techniques such as adaptive training strategies, few-shot domain adaptation, and reinforcement learning-based fine-tuning can enhance LLM performance in specialized fields by aligning model predictions with domain knowledge and linguistic conventions.

In summary, addressing domain-specific sentiment variability in LLMs requires a combination of domain-adapted training, hybrid modeling strategies, and structured fine-tuning approaches. Future research should focus on developing context-aware LLM architectures that can dynamically adjust sentiment predictions based on domain-specific linguistic cues, ensuring greater stability, accuracy, and interpretability in specialized sentiment classification tasks.

 \item \textit{Model scale and bias sentiment labels.}

 MVP in LLMs is significantly shaped by the interplay of model scale and bias in sentiment labels, both of which contribute to prediction instability and classification inconsistencies. While larger LLMs typically achieve higher performance on NLP tasks due to their ability to capture intricate language patterns, they are also more susceptible to overfitting heterogeneous pre-training data, thereby amplifying sentiment classification variability. This overfitting effect can lead to unstable sentiment predictions, particularly in cases involving nuanced, ambiguous, or context-dependent language. Furthermore, sentiment bias introduced during training distorts the output of the model, reinforcing systematic errors that affect interpretability and reproducibility in sentiment classification tasks.

Studies indicate that model size alone does not ensure stability in sentiment analysis. In \cite{ye2024benchmarking}, researchers find that larger LLMs tend to exhibit greater instability in sentiment classification compared to their smaller counterparts, particularly when dealing with subtleties such as sarcasm, implicit sentiment, or domain-specific emotional cues. This instability arises because larger models incorporate a broader range of linguistic patterns, making them more sensitive to slight variations in input phrasing or context shifts. Furthermore, the study highlights that as the complexity of the model increases, the variance in prediction scores also increases, reinforcing the trade-off between the scale of the model and the stability of the classification.

Beyond scale, bias in sentiment labels plays a crucial role in sentiment variability. \cite{shorinwa2024survey} highlights how bias in training data influences sentiment classification, leading to unpredictable changes in the interpretation of sentiment polarity. The study demonstrates that if a data set disproportionately represents certain sentiment categories or linguistic styles, the resulting LLM inherits and amplifies these biases, thereby producing skewed sentiment predictions. This phenomenon is particularly problematic in high-stakes applications such as financial sentiment analysis, political discourse evaluation, and brand reputation monitoring, where unintended biases can lead to systematic misinterpretation of sentiment trends.

Addressing these scale-related and bias-induced challenges requires a multifaceted mitigation strategy. First, bias-aware training methodologies should be integrated to ensure a balanced representation of sentiment in various linguistic styles and contexts. Second, uncertainty quantification techniques, such as Bayesian modeling and confidence calibration, can be used to estimate prediction variance and improve model robustness in sentiment classification. Finally, domain-specific fine-tuning and adversarial debiasing approaches can reduce susceptibility to biases datasets, ensuring that sentiment predictions remain consistent and interpretable regardless of variations in input phrasing or model scale.

In conclusion, while larger LLMs offer advanced linguistic comprehension, their susceptibility to overfitting and dataset-induced biases can contribute significantly to MVP. Future research must focus on developing scale-sensitive training techniques, implementing debiasing mechanisms, and integrating stability-enhancing methodologies to improve the reliability of sentiment classification in real-world applications.

 \item \textit{ Reinforcement learning fine-tuning and Reinforcement Learning  from Human Feedback.}
 
 MVP is significantly influenced by Reinforcement Learning Fine-Tuning, which introduces shifts in post-training model predictions. Reinforcement Learning from Human Feedback (RLHF) methdos are widely used to align LLM behavior with human preferences, ensuring that models generate ethically sound, coherent, and contextually appropriate responses. However, while RLHF improves alignment, it also introduces new challenges in sentiment analysis, as fine-tuning adjustments can make LLM outputs less predictable, leading to inconsistencies across sentiment classifications. If alignment updates are not properly calibrated, LLMs might develop biases or unpredictable shifts in sentiment prediction over time, leading to variability in model output for identical inputs.

Recent research highlights how alignment-induced changes introduce unpredictability in sentiment classification. In \cite{shorinwa2024survey}, it is shown that RLHF models tend to shift sentiment predictions unpredictably, as the fine-tuning process modifies model behavior based on subjective human feedback, leading to inconsistent sentiment classifications across different prompts and contexts. Similarly, \cite{beigi2024rethinking} discusses how post-training alignment mechanisms, such as safety filters, ethical constraints, and reinforcement objectives, can unintentionally distort sentiment interpretations, sometimes causing models to overcorrect or suppress certain sentiment polarities. These findings highlight a critical trade-off between model alignment and predictive stability, emphasizing the need for robust calibration techniques to maintain consistency in sentiment classification without introducing systematic distortions.

Beyond alignment challenges, fine-tuning itself introduces additional sources of model variability. Small variations in fine-tuning configurations, including random seed initialization, learning rates, training data variations, and slight modifications in hyperparameters, can lead to fine-tuning multiplicity, where multiple equally well-performing models generate conflicting sentiment classifications for the same input. This phenomenon is formalized in \cite{hamman2024quantifying}, which introduces a prediction consistency measure that demonstrates that different fine-tuned versions of the same base model can significantly diverge in the sentiment classification results due to subtle differences in training conditions. Such inconsistencies undermine the reliability of sentiment models, raising concerns about their stability, robustness, and reproducibility in high-stakes applications.

\item \textit{Human-AI interaction biases and adaptation challenges.}

One of the key contributors to MVP in sentiment analysis is the influence of human biases during interactions with LLMs. Human users inherently interact with LLMs in a subjective way and thus can introduce inconsistencies in the results of sentiment classification. These biases arise from cognitive tendencies such as automation bias and algorithm aversion, both of which influence how users interpret and rely on AI-generated output.

\begin{itemize}

\item Automation bias occurs when users overtrust AI-generated sentiment assessments, accepting outputs without critical scrutiny. This overreliance can reinforce systematic inconsistencies in model predictions, especially in ambiguous or context-sensitive sentiment classification tasks.

\item In contrast, algorithm aversion occurs when users develop skepticism toward AI models after experiencing errors or unexpected sentiment output. This can lead to unpredictable interactions, where some users override model decisions even when AI-generated sentiment assessments are accurate, reducing reproducibility and stability in AI-assisted decision-making processes.

\end{itemize}

As highlighted in \cite{passerini2025fostering}, human users exhibit distinct patterns of adaptation when engaging with LLMs, often reinforcing biased interpretations. Users who consistently trust AI-generated sentiment scores may unknowingly amplify model biases, embedding systematic distortions into the analysis pipeline. In contrast, users who frequently override AI decisions introduce variability by resisting the model output, creating instability in the consistency of sentiment assessment. This interaction-dependent bias raises concerns in critical applications, such as financial market sentiment analysis, healthcare sentiment evaluation, and policy-oriented opinion mining, where stable and unbiased sentiment predictions are essential for sound decision-making.

\item \textit{Bias in sentiment classification between LLM variants and multimodal sentiment conflicts.} 

 MVP in sentiment classification is exacerbated by bias inconsistencies between LLM versions and multimodal sentiment conflicts. Different iterations of LLMs, such as GPT-3.5 and GPT-4, frequently produce systematically different sentiment scores for the same input, highlighting discrepancies rooted in variations in their architectures, training methodologies, and fine-tuning approaches. These variations can result in drastic shifts in sentiment interpretation, making sentiment classification unpredictable across different versions of the LLM. Some models tend to exhibit inherent sentiment biases, systematically leaning toward positive or negative polarity, further contributing to instability in sentiment analysis.

Beyond text-based inconsistencies, multimodal sentiment conflicts arise when LLMs process and integrate information from multiple sources, such as text, images, and audio. Since LLMs primarily rely on textual context, their sentiment predictions often fail to synchronize with non-textual cues. For example, a neutral or slightly negative text statement might be accompanied by an image expressing a positive emotion, leading to conflicting sentiment classifications between modalities. Such misalignment presents significant challenges, particularly in applications such as social media sentiment tracking, customer experience analytics, and automated content moderation, where accurate sentiment extraction from multimodal inputs is critical.

 In \cite{krugmann2024-SAintheageofGenAI}, the authors found that GPT-3.5 exhibits a positive sentiment bias, while GPT-4 tends to be more neutral or negative, highlighting the inherent variability in sentiment classification between model versions. Similarly, \cite{zhang2023sentimentanalysiseralarge} identifies bias-driven instability, showing that different LLM architectures produce inconsistent sentiment predictions due to variations in training data distributions and fine-tuning methodologies. Furthermore, \cite{yang2024large} reveals alignment issues in multimodal AI, where discrepancies between text, image, and audio input create sentiment misclassification and conflicting interpretations, further exacerbating model inconsistency in sentiment analysis. These findings emphasize the need for bias mitigation strategies, standardized alignment protocols, and multimodal calibration techniques to improve stability in LLM-based sentiment classification.

\item \textit{Lack of calibration in confidence scores.}  

One of the critical factors contributing to MVP in LLM-based sentiment analysis is the lack of proper calibration of confidence scores. Confidence calibration refers to the alignment between the predicted confidence level of a model and its actual accuracy. LLMs often overestimate their confidence in incorrect predictions while underestimating it in accurate ones, leading to a disconnect between their perceived certainty and real-world performance. This miscalibration is particularly problematic in sentiment classification, where erratic confidence levels may cause inconsistent sentiment assignments, ultimately compromising the reliability of the model in decision-making processes. Without proper calibration, LLMs can misrepresent their predictive confidence, resulting in unstable sentiment scores on inference runs and reducing trust in AI-driven sentiment analysis applications.

 Empirical evidence is found in \cite{xie2025an}, where the discussion reflects how poorly calibrated models produce highly variable sentiment predictions, as inconsistencies in confidence estimation lead to overconfident but incorrect classifications or fluctuating sentiment scores across inference runs. Similarly, \cite{beigi2024rethinking} highlights that LLMs often lack uncertainty-aware calibration mechanisms, emphasizing that temperature scaling, Bayesian confidence adjustments, and quantile-based methods can improve model stability in sentiment analysis.
 
 These studies reveal that uncalibrated confidence scores introduce variability in model outputs, particularly in subjective sentiment tasks. Calibration errors lead to low trustworthiness in AI-generated sentiment classifications, which requires the integration of uncertainty quantification frameworks to improve LLM reliability in high-stakes applications.

\item \textit{Evaluation metrics limitations and sentiment evaluation benchmark.} 

One of the fundamental challenges exacerbating MVP in LLM-based sentiment analysis is the limitation of existing evaluation metrics and sentiment benchmarks. Traditional accuracy-based metrics, such as precision, recall, and the F1 score, fail to capture the variability inherent in LLM-generated sentiment classifications, as they primarily assess static performance without accounting for prediction inconsistency between inference runs. Similarly, existing sentiment benchmarks often oversimplify sentiment classification, relying on rigid categorical labels such as positive, negative, or neutral, which do not adequately reflect the complexity of real-world sentiment expressions, including sarcasm, contextual sentiment shifts, and mixed emotions. These constraints lead to inconsistent model evaluations, where the same model may yield different performance results depending on the benchmark used, further compounding uncertainty in LLM performance assessments.

 Empirical evidences are found in the following two studies. In \cite{krugmann2024-SAintheageofGenAI}, the authors critique existing sentiment benchmarks, demonstrating that they often fail to capture subtle sentiment transitions and contextual dependencies, which are essential for accurate sentiment interpretation. In \cite{ye2024benchmarking}, the authors highlight leaderboard discrepancies, showing that LLMs ranked highly in one evaluation setting may perform poorly in another, emphasizing the need for more robust benchmarking methods that account for sentiment stability and prediction consistency.

 These findings underscore the need for improved benchmarking frameworks that incorporate uncertainty-aware metrics, stability assessments, and real-world sentiment variations to provide a more accurate evaluation of LLM performance. Standard accuracy metrics do not assess intramodel consistency, leading to fluctuating model rankings across different datasets. Uncertainty-aware evaluation frameworks that incorporate prediction confidence, sentiment stability metrics, and context-aware assessments are needed to accurately measure LLM performance.

 \item \textit{The black-box nature of LLM decision-making.} 
 
 One of the key challenges contributing to MVP in sentiment analysis is the black-box nature of LLMs, which limits transparency and interpretability. LLMs generate sentiment classifications through complex neural architectures and large-scale statistical modeling, making it difficult to trace how and why a particular prediction is made. This opacity is problematic because identical inputs can yield different outputs, and without clear interpretability, debugging inconsistencies and mitigating variability remain significant challenges. The inability to explain these variations hinders trust in AI-driven sentiment models, particularly in high-stakes applications such as finance, healthcare care and policy analysis.

A key source of interpretability challenges in LLMs is the variability introduced by the grouping mechanisms used in sentence-embedded representations. Different pooling techniques determine how token-level embeddings are aggregated into a single sentiment representation, leading to inconsistencies in sentiment classification. Mean-pooling averages token embeddings, producing stable but sometimes diluted sentiment representations by smoothing out extremes. Max-pooling, on the other hand, captures the strongest sentiment feature by selecting the highest activation per dimension, emphasizing distinct sentiment features but at the cost of higher variability in predictions. Weighted sum pooling, which dynamically adjusts token importance based on learned weights, improves classification accuracy but increases interpretability challenges, as the influence of specific tokens is difficult to trace \cite{zhang2023sentimentanalysiseralarge}.
 
The pooling mechanisms directly affect the variability and interpretability of sentiment. In \cite{xing2024comparative}, the authors show that the sentiment classification outcomes fluctuate significantly depending on the grouping method used, highlighting how subtle changes in the pooling selection can alter model predictions. Similarly, \cite{beigi2024rethinking} underscores that the lack of interpretability in LLM intensifies the variability, making it difficult to diagnose sentiment inconsistencies. Furthermore, \cite{van2025advantages} contrasts LLM-based sentiment classification with lexicon-based approaches, demonstrating that lexicon models offer greater transparency and stability by relying on explicit sentiment word mappings rather than opaque neural representations. This suggests that hybrid models integrating lexicon-based and LLM-based approaches may offer a balance between accuracy and interpretability.

A related issue is overfitting to certain sentiment patterns due to pooling biases. In multimodal sentiment tasks, for example, weighted sum grouping can misallocate importance to sentimentally neutral words, distorting the final classification. In contrast, maximum pooling can amplify noise in sentiment classification, as it over prioritizes extreme words, leading to erratic output in cases where sentiment is ambiguous. These findings highlight the need for explanation-driven pooling selection methods, ensuring that LLMs prioritize stability and interpretability alongside accuracy.

To address the black-box problem, integrating XAI techniques is essential. Methods such as SHAP (SHapley Additive Explanations) and LIME (Local Interpretable Model-Agnostic Explanations) have been applied to sentiment analysis to clarify the role of individual words in influencing sentiment classifications. Attention visualization techniques have also been used to map out which words contribute the most to sentiment decisions, offering a clearer view of how sentiment shifts occur across inference runs.

Another promising direction is the development of structured pooling calibration techniques that reduce the interpretability-accuracy trade-off. Research in \cite{xing2024comparative} suggests that hybrid pooling methods ---combining mean and weighted sum pooling--- can achieve greater consistency while preserving contextual depth, making sentiment predictions robust and interpretable. Additionally, confidence calibration strategies, such as temperature scaling and Bayesian uncertainty modeling, can help align LLM predictions with actual model confidence, improving reliability and mitigating unpredictability in sentiment classification.

In summary, the black-box nature of LLM decision-making remains a central challenge in sentiment analysis variability. However, explainability-driven techniques, optimized pooling strategies, and interpretability-aware hybrid models offer practical solutions to improve transparency and stability. By integrating these approaches, future sentiment analysis models can minimize inconsistencies, enhance trust, and ensure that LLM-based sentiment classification remains accurate and interpretable.

\end{enumerate}

\section{Reflection and analysis on the importance of explainability for LLMs}
\label{sec:Exp}

The last point highlighted the issue of explainability and interpretability. In this section, attention is given to this problem as a fundamental problem for LLMs for user understanding and analysis.

Explainability \cite{arrieta2020explainable,ali2023explainable,longo2024explainable} can be considered an essential component of artificial intelligence. The following definition, proposed in \cite{arrieta2020explainable}, considers the two fundamental elements when we discuss explanations: \emph{understanding} and \emph{audience}. 
\vspace{2mm}
\begin{itemize}
    \item[] Given an audience, an \textit{explainable AI} (XAI) is one that produces details or reasons to make its functioning clear or easy to understand.
\end{itemize}
\vspace{2mm}

Today we can find a vast literature with different technical approximations to produce explanations for different AI systems. This research area, widely known as Explainable AI (XAI), has grown exponentially in the last few years, leading to recent discussions on its maturity and challenges \cite{ali2023explainable,longo2024explainable}. A detailed reflection on the XAI can be found in \cite{herrera2025}.

The growing complexity of LLM has underscored the urgent need for effective explainability frameworks. As discussed by Herrera (2025) in \cite{herrera2025}, the shift to black-box models in recent years has raised critical concerns about transparency, interpretability, and ultimately trustworthiness of AI systems. Herrera emphasizes that the increasing reliance on advanced AI models demands a comprehensive approach to XAI, not only as a tool for clarifying internal mechanisms but also as an essential factor for fostering trust and informed human-AI interaction. This notion aligns strongly with the identified need for robust interpretability in sentiment analysis using LLMs, particularly given the high-stakes and potential consequences associated with variability in sentiment output. Thus, the insights from Herrera’s reflections highlight the imperative to develop contextually grounded user-oriented XAI approaches capable of demystifying AI outputs and ensuring reliable decision-making across various critical applications.

Rapid adoption and integration of LLMs across diverse sectors underscores their transformative potential. However, despite their impressive capabilities in natural language processing, these models inherently function as \textit{"black boxes",} obscuring the decision-making processes behind their outputs \cite{zhao2024explainability}. This opacity presents critical challenges, particularly concerning transparency, reliability, and ethical responsibility, that require the rigorous exploration and development of XAI methodologies.

One of the fundamental motivations for explainability in LLMs is trust building. As Luo and Specia (2024) highlight \cite{luo2024understanding}, the confidence of end-users in AI-driven systems is significantly dependent on their ability to understand the reasoning behind specific predictions or classifications. Without clear explanations, stakeholders cannot reliably calibrate model performance, leading to the potential misuse or rejection of valuable AI tools. This trust factor is especially critical in sensitive domains like healthcare, finance, and legal decision making, where misunderstood or inaccurate model predictions can have severe consequences \cite{zhao2024explainability}.

Furthermore, as the study \cite{barman2024beyond} argues, focusing solely on transparency in the abstract may not adequately address the practical needs of diverse user groups. They emphasize the need for explainability methods that not only clarify why a model made a particular decision but also provide actionable, contextual guidance for users. Effective explainability, therefore, should move beyond mere transparency towards enabling practical, contextualized understanding that facilitates responsible AI use. This perspective suggests a shift from purely technical explanations toward pragmatic guidelines tailored to specific use cases and user proficiency levels.

Addressing explainability in LLMs, \cite{zhao2024explainability} introduces a comprehensive taxonomy of techniques, categorizing them into local and global explanations based on their explanatory objectives. Local explanations, such as feature attribution, attention visualization and counterfactual explanations, elucidate the reasoning of the model for specific predictions, directly supporting user trust by providing concrete justifications for outputs. These techniques enable users to precisely understand which inputs or features most significantly influence individual decisions, making it possible to validate or challenge predictions on a case-by-case basis. In contrast, global explanations, which encompass approaches such as classifier investigation, mechanistic interpretability, and representation analysis, assist researchers in comprehending the overarching behaviors and structural properties of the model. These methods provide insight into internal mechanisms, hidden biases, and general knowledge encoded within the models, thus playing a crucial role in debugging, systematic model improvements, and identifying vulnerabilities or systemic issues such as embedded societal biases or the tendency of models to generate misleading information. This taxonomy emphasizes the complementary nature of local and global methods, highlighting the necessity of integrating multiple explanation types to build comprehensive, trustworthy, and interpretable LLMs.

However, current explainability methods still face significant challenges, especially due to the unprecedented scale and complexity of modern transformer-based LLMs such as GPT-4 or LLaMA. As emphasized by \cite{luo2024understanding}, traditional techniques such as SHAP or LIME, while valuable, often struggle with computational scalability when applied to models with billions of parameters. Moreover, existing explainability metrics frequently fall short in capturing nuances in model behaviors such as in-context learning and chain-of-thought reasoning, highlighting the need for novel, efficient, and scalable explanatory methods tailored explicitly to large-scale generative models.

In addition, explainability plays a crucial role in mitigating the ethical and social risks associated with LLM. In \cite{zhao2024explainability}, they point out that opaque decision-making processes in models often lead to unintended biases, harmful content generation, and hallucinations, outcomes that pose substantial ethical and social implications. Robust explainability frameworks enable proactive identification and mitigation of such risks, fostering ethical alignment and responsible deployment of these powerful technologies.

In summary, advancing explainability in LLMs is essential not only to enhance user trust and model reliability but also to ensure ethical and responsible use of AI technologies in society. In the future, the development of more sophisticated, context-sensitive, and user-oriented XAI frameworks, as advocated in \cite{barman2024beyond}, is crucial. Future research and practical guidelines should aim to bridge theoretical understandings and empirical methodologies with real-world user-centric applications, ensuring that the profound capabilities of LLMs can be channeled ethically, responsibly, and effectively.

Integrating XAI techniques into LLM-based sentiment analysis is crucial to enhance model transparency and reduce uncertainty. By providing clear insights into how models arrive at their predictions, XAI facilitates the identification and mitigation of inconsistencies and biases inherent in LLMs. Techniques such as attention mechanisms and visualization tools can highlight which parts of the input text most influence sentiment predictions, enabling users to understand and trust the model's decision-making process. For example, employing attention-based explanation methods can reveal how specific words or phrases contribute to the overall sentiment classification, thereby offering a more interpretable and reliable analysis. Furthermore, approaches grounded in linguistic theory, such as construction grammar, can further clarify how LLMs internalize complex linguistic patterns and meanings, thus providing deeper explanatory insights into model behavior and potential misunderstandings \cite{weissweiler2023explaining}. By adopting these combined XAI strategies, stakeholders can achieve a deeper understanding of model behavior, leading to more consistent and trustworthy sentiment analysis results \cite{mabokela2024explainable, weissweiler2023explaining}.

As highlighted by \cite{da2025understanding}, the reasoning processes in LLMs are not always stable, with different paths leading to the same or different final conclusions depending on how the model structures its logical dependencies. This structural uncertainty, when applied to sentiment analysis, can result in divergent sentiment scores for the same input. Addressing this challenge requires adopting structured methods for explanation-based uncertainty quantification, such as reasoning topology modeling, to systematically capture and mitigate variability in sentiment interpretation.

In summary, the exploration of explainability plays a vital role in addressing the uncertainty and variability inherent in LLM-based sentiment analysis. Given that LLM sentiment predictions are particularly susceptible to variability driven by stochastic inference, prompt sensitivity, and training biases, it is crucial to incorporate robust and interpretable explanation methods. Effective explainability not only provides clarity on how and why sentiment classifications vary, but also establishes trust and facilitates user validation of AI-generated sentiments. Therefore, aligning comprehensive XAI frameworks with methods for uncertainty quantification and variability mitigation forms an essential foundation for improving the stability, reliability, and trustworthiness of LLM-based sentiment analysis systems, directly addressing the central themes of this perspective study.

\section{Challenges for Model Variability Problem in LLM-based sentiment analysis}
\label{sec:Challenges}

The evaluation of LLM-based sentiment analysis faces significant challenges due to MVP, which affects reliability, interpretability, and applicability across domains. This section identifies and explores twelve important challenges that we have found in a literature review that raise prospects for addressing the problem with solutions that reduce variability. 

\begin{enumerate}
    \item \textit{Lack of standardized and stability-aware benchmarking frameworks.} 
    
    The evaluation of sentiment analysis models is based primarily on traditional performance metrics such as precision, precision, recall, and the F1 score. However, these metrics are insufficient to assess variability in LLM-based sentiment analysis, as they do not account for fluctuations in model output over multiple inference runs. Although standard sentiment benchmarks exist, they often fail to measure stability and consistency, making it difficult to determine whether an LLM can reliably classify sentiment in different contexts and prompts.

    One of the key issues is that models that rank in one benchmark may perform poorly in another, as highlighted in \cite{ye2024benchmarking}, demonstrating benchmarking inconsistencies in sentiment analysis. Furthermore, traditional evaluation metrics do not consider how sentiment classification changes over time due to model updates, prompt variations, or data set changes. This gap in the evaluation leads to a misleading perception of the reliability of the model, which prevents a proper assessment of the stability of the prediction of sentiments. 
    
    The instability of sentiment classification outputs in LLMs is exacerbated by the model's sensitivity to prompt construction and hyperparameter settings. Recent research demonstrates that sentiment scores can vary significantly due to minor prompt rewording or changes in decoding parameters, reinforcing the need for methods to improve stability and reliability in sentiment classification \cite{loya2023exploring}.\\

  \textit{Potential solution.}  A potential solution is to introduce stability-aware benchmarking frameworks that incorporate uncertainty-aware evaluation metrics, such as confidence scores based on entropy and stability indices. Furthermore, cross-benchmark validation using multiple domain-specific datasets could improve sentiment classification robustness, ensuring that evaluations reflect real-world performance variations rather than overfitting to specific benchmarks.\\

 \item \textit{Sensitivity to prompt variability and input reframing.} 
 
 LLMs are highly sensitive to variations in prompt phrasing, which means that small changes in input wording can lead to significantly different sentiment predictions. This issue makes it difficult to replicate the results of sentiment classification consistently, reducing trust in LLM-based sentiment analysis applications.

    Studies such as \cite{zhang2023sentimentanalysiseralarge} and \cite{krugmann2024-SAintheageofGenAI} reveal that GPT-3.5 and GPT-4 generate different sentiment scores based on prompt specificity, even when the sentiment meaning remains unchanged. Furthermore, the lack of standardized prompt design guidelines makes it difficult for researchers to evaluate whether the sentiment output of a model is a result of genuine contextual understanding or a by-product of prompt sensitivity.

Recent work by \cite{reveilhac2024chatgpt} shows that LLMs such as ChatGPT exhibit notable shifts in decision-making behavior based on linguistic, cultural, and contextual cues embedded within the prompts. The study highlights how model outputs fluctuate depending on the ideological framing of political prompts and the language in which queries are presented. This variability aligns with our findings that sentiment analysis outputs are highly prompt-sensitive, requiring standardized prompt engineering strategies to mitigate inconsistencies.\\

\textit{Potential solution.}  Given the substantial impact of prompt sensitivity on the variability of sentiment analysis, it is essential to develop systematic techniques to mitigate its effects. Key strategies include:

\begin{itemize}
    \item Standardization of prompt design to reduce linguistic variability and improve consistency in model responses.
    \item  Implement prompt optimization frameworks that ensure that LLMs generate sentiment predictions with minimal deviation across inference runs.
    \item  Integrating uncertainty-aware modeling to detect when model predictions are likely to fluctuate due to prompt variations.
    \item Using ensemble-based sentiment evaluation methods, where multiple prompt structures are tested to derive more robust and consensus-driven sentiment scores.
\end{itemize}

By addressing prompt sensitivity through structured input standardization and explainability-driven interventions, LLM-based sentiment analysis can achieve greater stability, reliability, and reproducibility, ensuring its effective deployment in high-stakes AI-driven decision-making environments.\\

 \item \textit{Epistemic and aleatoric uncertainty in the interpretability of the model.}
 
LLMs suffer from two major types of uncertainty: epistemic uncertainty, which arises from knowledge limitations within the model, and aleatoric uncertainty, which is caused by inherent noise and ambiguity in training data. These uncertainties make it difficult for LLMs to produce stable sentiment classifications, leading to inconsistencies in sentiment predictions.

As discussed in \cite{shorinwa2024survey}, epistemic uncertainty is the result of incomplete training data, causing LLMs to misinterpret sentiment in unseen or ambiguous contexts. Furthermore, aleatoric uncertainty, as highlighted in \cite{beigi2024rethinking}, arises due to human annotation errors, slang, sarcasm, and domain shifts, leading to inconsistent sentiment classifications. This dual source uncertainty problem reduces the reliability of sentiment analysis models. In \cite{reveilhac2024chatgpt}, the authors emphasize how epistemic uncertainty affects decision making in LLM-powered voting systems, and \cite{passerini2025fostering} explores how mutual adaptation between humans and LLM can reduce or amplify epistemic uncertainty. 

The association of aleatoric and epistemic uncertainty with the studies analyzed reveals that both sources of uncertainty are the main contributors to LLM variability in sentiment analysis. Aleatoric uncertainty arises from intrinsic linguistic ambiguities, while epistemic uncertainty arises from knowledge limitations within the model itself. \\

\textit{Potential solution.}  Effective mitigation requires a combination of context-aware sentiment embeddings, structured fine-tuning methodologies, confidence calibration frameworks and explainability-driven feedback mechanisms. By addressing these uncertainties, we can significantly improve the robustness, reliability, and interpretability of sentiment classification in LLM-based systems.

To mitigate uncertainty in sentiment analysis, models should be trained with uncertainty-aware learning techniques, such as Bayesian deep learning or Monte Carlo dropout, which quantify and account for confidence levels in sentiment predictions. Additionally, incorporating explainability methods (e.g., attention visualization) can help researchers diagnose whether sentiment shifts are due to knowledge gaps or data-driven noise, improving interpretability and reliability.\\

\item \textit{RLHF-induced variability.}

RLHF and fine-tuning methodologies play a crucial role in aligning LLM sentiment predictions, but they also introduce new sources of variability that must be carefully managed. Future research should focus on developing structured RLHF calibration frameworks, stability-aware fine-tuning techniques, and ensemble-based evaluation strategies to ensure consistency in LLM-based sentiment classification tasks. By addressing these alignment-induced inconsistencies, sentiment analysis models can achieve greater reliability, reproducibility, and robustness, enabling more effective deployment in real-world decision-making environments. \\

\textit{Potential solution.}Addressing the variability introduced by RLHF and the fine-tuning multiplicity requires the implementation of structured calibration techniques and stability-aware optimization frameworks. Key strategies include:

\begin{itemize}

\item  Confidence-aware RLHF adjustments: Fine-tuning alignment strategies should incorporate confidence estimation techniques that assess the impact of reinforcement learning updates on prediction consistency before deployment.
\item  Fine-tuning stability protocols: Employing stability-driven retraining techniques, such as ensemble fine-tuning, iterative feedback alignment, and controlled learning rate decay, can reduce fluctuations in sentiment classification.
\item  Ensemble-based consistency Evaluation: Using multiple fine-tuned model checkpoints and aggregating predictions through voting mechanisms can increase robustness and reduce the influence of alignment-induced shifts.
\item  Uncertainty quantification techniques: Implementing quantile-based calibration, epistemic uncertainty estimation, and Monte Carlo dropout methods can help quantify variability in fine-tuned sentiment models, ensuring more reliable output.\\

\end{itemize}

\item \textit{Sensitivity to model updates and fine-tuning variability.}

One of the major challenges in ensuring stability in sentiment analysis is the variability introduced by iterative model updates and fine-tuning strategies. Fine-tuned LLMs, even when trained on similar datasets with minor modifications, may produce contradictory sentiment predictions for identical inputs, introducing inconsistencies in high-stakes applications like finance, healthcare, or customer feedback analysis. This issue is exacerbated by the need for frequent model retraining due to data drift, evolving user language, and regulatory constraints such as GDPR's 'right to be forgotten', which requires data removal and potential model retraining.

Recent work by \cite{hamman2024quantifying} systematically examines the fine-tuning multiplicity, demonstrating that models fine-tuned under slightly different conditions (e.g. random seed initialization, additional training samples) can exhibit arbitrary sentiment classification results. Their proposed prediction consistency metric quantifies a model's susceptibility to fine-tuning variability and offers a probabilistic measure of prediction robustness. Addressing this challenge requires developing stability-sensitive fine-tuning protocols, uncertainty-sensitive retraining strategies, and adaptive sentiment calibration techniques that mitigate inconsistencies arising from iterative model updates.\\

\textit{Potential solution.} Addressing the instability caused by the variability in fine-tuning requires a combination of stability-aware training protocols, uncertainty quantification techniques, and adaptive calibration strategies. One approach is to implement ensemble fine-tuning, where multiple instances of a fine-tuned model are trained with different initializations and hyperparameters, and their outputs are aggregated using consensus mechanisms to enhance prediction robustness. In addition, regularization techniques, such as variance penalization during training, can help reduce divergence among fine-tuned models. Another promising method is progressive fine-tuning, where model updates are applied in smaller, controlled increments to minimize abrupt shifts in sentiment classification behavior.  Finally, continuous learning strategies that incorporate past training checkpoints while adapting to new data distributions can improve the consistency of model update, reducing the likelihood of erratic sentiment classification shifts over time.\\

\item \textit{Reproducibility and stability in sentiment analysis.}

One of the biggest challenges in the deployment of LLMs for sentiment analysis is ensuring reproducibility and stability. The lack of deterministic behavior in LLM introduces output fluctuations that affect reliability in real-world decision-making applications. Stability and reproducibility remain key unresolved challenges in LLM-based sentiment analysis.

As highlighted in \cite{atil2024llm}, LLMs often generate inconsistent responses even when the same input is provided under supposedly deterministic settings (e.g., temperature=0, fixed seeds, identical prompts). This behavior significantly affects use cases that demand repeatability, such as financial sentiment analysis or healthcare AI applications. This unpredictability arises from temperature variations, randomness in the decoding, and latent model instabilities, which can cause the same sentiment query to be interpreted differently between runs. The study further notes that longer output sequences correlate negatively with stability, meaning that models generating verbose explanations tend to exhibit even greater variability.\\

\textit{Potential solution.} To address this challenge, it is crucial to establish new standardized benchmarking protocols that assess model consistency over repeated runs. In addition, integrating ensemble techniques or model voting mechanisms could improve decision stability by averaging individual model fluctuations. Furthermore, fine-tuning strategies that explicitly optimize for deterministic behavior and constrain variance using regularization techniques could mitigate these inconsistencies, fostering a more reliable foundation for LLM-driven sentiment analysis.

On the other hand, regarding inference stochasticity in general, and the temperature variations in particular, we can pay attention to different actions.  Among these to mitigate the temperature-induced variability, several strategies can be considered:

\begin{itemize}
   \item Use of low-temperature sampling ($T \approx 0$): For tasks requiring deterministic or audit-ready outputs, setting the temperature to near zero can reduce stochasticity. However, this may also increase the exposure to alignment flaws or high-confidence errors.
  \item Multi-sample aggregation: Sampling multiple outputs at varied temperatures and applying majority-vote or confidence-based aggregation can smooth stochastic spikes.
  \item Temperature calibration curves: Tracking output divergence across a range of temperature settings allows the identification of robust operating zones for a given task or domain.

\end{itemize}

\item \textit{Human-LLM feedback loop and confirmation bias.}

Another significant challenge in LLM-based sentiment analysis is the reinforcement of feedback loops that arise from repeated human-LLM interactions. This issue stems from confirmation bias, where users unintentionally reinforce preexisting beliefs by influencing how LLMs generate responses. Rather than acting as neutral sentiment classifiers, models can gradually align with user expectations, amplifying subjective biases rather than maintaining objective sentiment analysis.

In \cite{passerini2025fostering}, researchers discuss how mutual adaptation between humans and LLMs can either enhance decision-making synergy or exacerbate cognitive biases. In sentiment analysis tasks, this manifests itself when users consistently interact with an LLM in a way that skews its responses toward a specific sentiment polarity. Over time, models may reinforce the user's preferred sentiment interpretations, leading to skewed sentiment scores that lack objective grounding. This self-reinforcing loop compromises neutrality, particularly in sensitive applications such as public opinion analysis, market research, and policy evaluation. Another significant challenge is the way human feedback loops influence LLM behavior, leading to reinforcement of biased sentiment interpretations.

Human-LLM interaction can lead to a self-reinforcing feedback loop, where model responses adapt to user expectations rather than maintaining an objective stance. As users engage with sentiment analysis systems over time, the likelihood of confirmation bias increases, where the model prioritizes responses that align with previously accepted sentiment patterns, rather than evaluating text based on an independent, neutral linguistic framework. This issue is exacerbated by the iterative fine-tuning of LLMs based on user interactions, as models continuously learn from their own outputs and user preferences, further embedding biases into future predictions.\\

\textit{Potential solution.} To mitigate this problem, adaptive bias correction techniques and trust-based LLM calibration are essential. Possible solutions include:
\begin{itemize}
\item Counter-bias mechanisms, where models periodically introduce neutral or opposing perspectives to break reinforcing feedback cycles.
\item	Diversity-driven sentiment prompts, encouraging users to interact with varied perspectives, preventing model drift toward biased sentiment outputs.
\item Interactive explainability features, allowing users to assess the reasoning behind sentiment predictions, thereby fostering more balanced decision-making.
\item  Adaptive AI-human collaboration frameworks, where LLMs dynamically adjust response strategies based on user interaction patterns, ensuring more stable and objective sentiment analysis outputs.

\end{itemize}

By integrating bias-aware interaction models, human-AI collaboration can be optimized, ensuring higher reliability, objectivity, and consistency in sentiment classification.\\

\item \textit{Bias-induced variability and domain adaptation issues.} 

LLMs inherit biases from their training data, leading to systematic sentiment variations when analyzing content from different demographics, industries, or social groups. This bias-induced variability can cause sentiment misclassification, making it difficult for models to maintain consistent results across different domains.

A major concern in LLM-based sentiment analysis is bias and inconsistency in sentiment aggregation, where models generate contradictory sentiment scores for the same input text based on prompt structure, response aggregation methods, or temperature settings. Research on LLM-based voting mechanisms suggests that aggregation inconsistencies are particularly pronounced in multiwinner voting settings, where the ordering of options and voting methodologies affects decision-making outputs \cite{yang2024llm}. These findings indicate that similar aggregation challenges exist in sentiment analysis tasks, where ranking-based sentiment scoring versus binary sentiment classification can produce conflicting results. This highlights the importance of structured sentiment evaluation strategies, where results are normalized across prompts and calibrated to mitigate sensitivity to order effects.

 As shown in \cite{krugmann2024-SAintheageofGenAI}, GPT-3.5 exhibits positive sentiment bias, while GPT-4 leans more neutral or negative, demonstrating how sentiment polarity can vary between model versions. Furthermore, \cite{van2025advantages} highlights that LLMs struggle with domain-specific sentiment tasks, making them unstable when applied to specialized fields like finance or healthcare. These findings emphasize that bias-related variability affects the reproducibility of sentiment classification. 

As demonstrated by \cite{reveilhac2024chatgpt}, differences in the versions of the LLM model can lead to inconsistent decision-making outcomes, even for politically relevant tasks such as voting simulation. The study found that the ideological positioning of ChatGPT changes between GPT-3.5 and GPT-4, with version-dependent biases that affect the stance taken on issues of direct democracy. In sentiment analysis, such model drift raises concerns about reproducibility, making longitudinal stability evaluations essential for benchmarking consistency in AI-generated sentiment predictions.\\

\textit{Potential solution.} A key strategy to mitigate bias-induced variability is to incorporate fairness-sensitive training methods, such as bias correction techniques and adversarial debiasing. Additionally, fine-tuning LLMs in domain-specific sentiment datasets can improve stability and prevent sentiment misclassification when models are applied to specialized contexts.\\

\item \textit{Achieving consensus through ensemble approaches among LLM in sentiment analysis}

One promising approach to mitigating MVP in sentiment analysis involves leveraging ensemble methods, where multiple LLMs independently evaluate the same textual input, subsequently seeking a consensus or aggregated decision. Two studies \cite{agrawal2024ensemw2s,abburi2023generative} demonstrate that combining multiple models can enhance robustness and predictive stability. However, effectively achieving consensus among independently operating LLMs for sentiment classification remains challenging, as individual models may differ significantly in their predictions due to inherent architectural differences, different training procedures, and varied inference strategies.

A primary issue is managing divergent outputs from multiple models, as ensemble members can provide contrasting sentiment scores due to differences in training data, biases, prompt sensitivity, and stochasticity in token generation.  In \cite{agrawal2024ensemw2s}, it is highlighted that ensembles can sometimes amplify rather than reduce variance if not appropriately managed, particularly when member models differ substantially in reliability or calibration. 

In \cite{abburi2023generative}, additional complexities are emphasized in aligning and interpreting the probabilistic predictions generated by each member of the ensemble, creating difficulties in determining a classification of unified feelings. Without a structured mechanism for aggregation and reconciliation, ensemble approaches risk exacerbating interpretability challenges, increasing computational costs, and potentially diminishing user trust.

A significant challenge in sentiment analysis is to ensure consistency when aggregating the outputs of multiple LLMs to derive a final sentiment classification. This issue becomes particularly relevant in LLM-based ensemble voting mechanisms, where multiple models contribute to a collective decision. In \cite{jarrett2025language}, the authors explore how digital representatives can effectively simulate human decision making within collective settings, providing insights into how LLMs can be aligned for more structured decision aggregation. However, as our study highlights, variability in individual LLM predictions complicates the consensus-building process, as slight variations in model outputs can lead to significantly different aggregated sentiment classifications. Addressing this challenge requires the integration of consensus-driven voting algorithms, adaptive weighting schemes, and trust-based model selection mechanisms to enhance consistency across multi-LLM systems.\\

\textit{Potential solution.} To address these challenges, a structured and adaptive ensemble consensus strategy can be adopted. Specifically, methods such as weighted aggregation, confidence-based voting, or adaptive majority voting informed by uncertainty quantification metrics can help effectively reconcile divergent sentiment scores into a unified and trustworthy classification. Inspired by Agrawal et al. (2024), incorporating token-level weighting or boosting mechanisms can further enhance consensus stability. Furthermore, the integration of transparent and explainable aggregation mechanisms, such as visualization tools that explicitly illustrate how each model contributes to the final sentiment decision, would significantly improve interpretability and user trust. Such consensus-driven ensemble methods not only promise more stable and reliable sentiment predictions but also foster greater acceptance and confidence in LLM-based sentiment analysis outcomes.\\

\item \textit{Applying knowledge distillation to mitigate MVP in LLMs for sentiment analysis}.

LLMs have significantly advanced natural language processing tasks, including sentiment analysis. However, their substantial computational requirements and inherent model variability pose challenges for practical deployment. Knowledge distillation \cite{gu2024minillm,gu2024miniplm} ---a technique in which a smaller "student" model learns from a larger "teacher" model--- offers a potential solution to these issues. The primary objective is to maintain the performance of the original model while reducing its size and computational demands. However, effective application of knowledge distillation to LLMs, particularly in the context of sentiment analysis, presents specific challenges that need to be addressed.

Training Smaller Language Models (SLM) through knowledge distillation presents several challenges. Firstly, LLMs possess intricate architectures capable of capturing nuanced language patterns, making it difficult to transfer this sophisticated knowledge to smaller models without significant performance loss. Additionally, standard distillation methods may not be optimal for generative language models due to differences in output distributions between teacher and student models, leading to suboptimal performance in the distilled models. Moreover, LLMs often produce uncertain or ambiguous outputs, especially in tasks like sentiment analysis that involve nuanced expressions; effectively capturing and transferring this uncertainty during distillation is complex, but crucial for maintaining model reliability. Recent studies have explored methods to improve knowledge distillation for small language models, aiming to address these challenges and improve the efficiency and effectiveness of the distillation process \cite{yam2024teaching}. \\

\textit{Potential solution.} To enhance the performance and efficiency of SLM, researchers have explored the integration of knowledge distillation and fine-tuning techniques. For example, the MiniPLM framework \cite{gu2024miniplm} refines the distribution of training data using the teacher's insight, enabling the student model to achieve competitive performance with reduced complexity. Additionally, fine-tuning the distilled model on task-specific data allows it to adapt to particular nuances, further enhancing its effectiveness in applications like sentiment analysis. This combined approach not only maintains robust performance but also ensures that smaller models are more adaptable and efficient in real-world scenarios.\\

\item \textit{Ensuring consistency and robustness in crowd decision making through prompt-based LLM.} 

A significant challenge in leveraging LLMs for crowd decision-making involves ensuring consistency and robustness in sentiment classification outputs derived from structured prompt interactions. Despite the flexibility and effectiveness of prompt design strategies with models such as ChatGPT, these approaches are highly susceptible to variability, as small modifications in prompt wording, context framing, or inference parameters can result in substantial fluctuations in sentiment scores or classifications. The problem is exacerbated when opinions aggregated from texts from various users are sensitive to variations in language nuances, leading to uncertain or conflicting decisions \cite{Herrera-Poyatos2025large}.

Key issues associated with this challenge include the inherent sensitivity of LLMs to slight prompt modifications, causing unpredictable shifts in sentiment polarity assessments. As highlighted in \cite{Herrera-Poyatos2025large}, this sensitivity leads to inconsistencies that hinder stable decision-making processes, undermining the reliability of CDM systems. Furthermore, the complexity of aggregating multiple sentiment evaluations from various crowd sources magnifies these inconsistencies, complicating consensus formation, and potentially reducing trustworthiness. Furthermore, without adequate transparency and explainability mechanisms, stakeholders cannot readily identify the rationale behind divergent model outcomes, intensifying the uncertainty surrounding the final decisions.\\

\textit{Potential solution.} To develop structured frameworks and standardize and optimize prompt designs, minimizing variability through well-defined linguistic templates. Additionally, employing ensemble consensus methods, as suggested in recent literature, can significantly stabilize LLM-based sentiment predictions by aggregating outputs from multiple structured prompts or diverse LLM instances, smoothing out variations and enhancing robustness. Integrating uncertainty quantification metrics into the CDM workflow could further help assess the reliability of aggregated sentiment scores, allowing decision makers to explicitly evaluate confidence levels. Finally, coupling these methodological strategies with explainability tools, such as attention visualization or token attribution methods, would increase the transparency of decision-making processes, facilitating trust and acceptance among stakeholders in CDM contexts utilizing LLM.\\

 \item \textit{Model variability in the era of open-source LLM proliferation} 

Rapid adoption of open-weight models such as DeepSeek-R1 has fundamentally shifted the landscape of LLM research and deployment. No longer lagging behind their proprietary counterparts, open-source LLMs now offer state-of-the-art reasoning and instruction-following abilities, with transparent architectures and permissive licensing. These models are increasingly chosen for cloud integration, mass-market applications (e.g., Perplexity), and downstream fine-tuning. However, this shift introduces new and urgent challenges in terms of model variability, reproducibility, and trustworthiness.

Unlike proprietary models that are version-locked and centrally maintained, open-source models such as DeepSeek-R1 and Falcon are being adapted, compressed, and fine-tuned in thousands of independent forks. This leads to substantial behavioral drift, including variation in output under prompt rephrasing, inconsistent refusal behavior, and degraded factuality due to uneven fine-tuning practices. Compression techniques such as CompactifAI \cite{tomut2024compactifai}, although valuable for scalability, can amplify these variabilities unless systematically controlled.

This fragmentation makes it difficult for researchers and enterprise users to establish behavioral guarantees or audit model decisions. Addressing the MVP in open-source ecosystems is critical for a trustworthy deployment. This challenge requires the development of evaluation protocols and mitigation strategies to ensure stability, safety, and consistency, even across diverse implementations, compression states, and usage contexts. As open models begin to replace closed APIs in production systems, this issue will become increasingly central to the future of reliable open LLMs.

\textit{Potential solution.} Addressing model variability in the context of open source LLM proliferation requires a multifaceted strategy that combines methodological standardization, ensemble stabilization, and post-deployment monitoring. One promising direction is the development of benchmarked prompt engineering templates and shared inference protocols that reduce behavioral drift between implementations. In parallel, ensemble-based approaches, such as majority voting or confidence-weighted aggregation across multiple fine-tuned instances, can improve stability by smoothing out individual model fluctuations. Additionally, incorporating calibration and consistency checking techniques, such as response entropy tracking and top k divergence metrics, can help identify and correct unstable behaviors. It is also crucial to adopt reproducibility-aware practices, including model cards and traceable configuration logs, to document the provenance and training variations of open source forks. Lastly, open models should integrate lightweight explainability modules or behavioral validation tests (e.g., prompt response unit tests) during compression or fine-tuning, to proactively detect and mitigate variability. These strategies aim to preserve the flexibility of open source LLMs while introducing essential reliability layers for production- and research-grade applications.\\

\item \textit{Lack of explainability and trustworthiness in the output of sentimental polarity.}

As we have discussed in Section \ref{sec:Exp}, one of the central challenges in the analysis of sentiment with LLMs is their lack of transparency and explainability, which undermines trustworthiness and reliability. LLMs operate primarily as black-box systems, obscuring the reasoning behind sentiment polarity predictions and making it challenging for users to understand how conclusions are reached. Unlike lexicon-based methods, where explicit word-sentiment mappings allow transparent reasoning, LLMs rely on intricate neural network representations. This complexity impedes the ability of users and stakeholders, especially in sensitive applications, to trust sentiment predictions, as decisions based on unclear or inconsistent outputs pose substantial risks.

The opaque nature of LLMs creates significant hurdles for reproducibility and trustworthiness. Users cannot reliably interpret why particular sentiment scores are assigned, especially when these scores fluctuate over multiple inference runs, exacerbating the MVP. This lack of interpretability not only compromises the transparency of the model, but also undermines trust, as stakeholders cannot confidently understand or justify sentiment predictions in critical scenarios. Research such as \cite{van2025advantages} explicitly contrasts the interpretability benefits of lexicon-based approaches with the inherent unpredictability and lack of transparency of LLM-based sentiment analysis. Furthermore, as shown in \cite{beigi2024rethinking}, LLM sentiment scores frequently fluctuate with minor changes in model alignment methods or training data nuances, highlighting how inherent black-box characteristics amplify concerns about model trustworthiness and reliability.

One of the key challenges in ensuring the interpretability of LLM-based sentiment analysis is understanding how different interaction paradigms influence sentiment classification decisions. The taxonomy presented in \cite{gao2024taxonomy} categorizes the interaction modes that structure human-LLM exchanges, providing insights into how structured prompting, UI-enhanced reasoning, and agent-facilitated collaboration affect model transparency. These structured approaches can improve explainability by making LLM-generated sentiment scores more interpretable, reducing user uncertainty, and fostering greater trust in AI-driven sentiment analysis. In \cite{da2025understanding}, the study introduces a reasoning topology framework that decomposes LLM-generated explanations into structured components, allowing for a more precise evaluation of uncertainty. This approach is particularly relevant for sentiment analysis, where model variability can often arise from subtle differences in justification paths. Using structured uncertainty quantification, models can be designed to provide more consistent and interpretable sentiment predictions, reducing the overall impact of stochastic variability.\\

\textit{Potential solution.} To improve interpretability, reliability, and trustworthiness in sentiment analysis, the integration of XAI techniques is crucial. Methods such as SHAP (SHapley Additive Explanations) and LIME (Local Interpretable Model-Agnostic Explanations) can be utilized to visualize and quantify the contributions of input tokens, enabling a clearer tracing of sentiment decisions. Another effective approach involves creating hybrid sentiment analysis systems that integrate LLM's contextual awareness with transparent, lexicon-based sentiment rules. Furthermore, structured uncertainty quantification and calibration frameworks, such as confidence-based prediction intervals or ensemble models, can further enhance reliability, ensuring that users not only receive interpretable outputs but also consistently reliable sentiment classifications. By explicitly addressing interpretability and trustworthiness alongside reliability, sentiment analysis models become better aligned with stakeholder expectations, fostering greater acceptance and integration in critical decision-making workflows.

\end{enumerate}

The thirteen major challenges discussed, benchmarking limitations, prompt sensitivity, uncertainty in interpretation, bias-induced variability, and impact of the pooling mechanism, underscore why LLM-based sentiment analysis remains highly variable and difficult to evaluate consistently. In summary, addressing these challenges requires the following actions.

\begin{itemize}
    
    \item To improve stability-aware sentiment evaluation metrics and mitigation strategies.
    \item To standardize prompt design frameworks to reduce sensitivity.
    \item To improve interpretability by means of uncertainty-aware training techniques.
    \item To integrate bias mitigation strategies and domain-adaptive fine-tuning.
    \item To achieve consensus through ensemble approaches among LLMs in sentiment analysis.
    \item To achieve the reproducibility and stability in sentiment analysis
    \item To apply knowledge distillation to mitigate MVP in LLMs for Sentiment Analysis, to obtain SLMs. 
    \item To ensure consistency and robustness in CDM via prompt-based LLMs
    \item To enhance interpretability, reliability, and trustworthiness in sentiment analysis with the  integration of XAI. 
    \item To promote open source development with reproducibility-aware practices, including shared inference protocols, model cards, and behavioral validation during compression and fine-tuning.
\end{itemize}

By implementing these solutions, the reliability and trustworthiness of LLM-based sentiment analysis can be significantly improved, paving the way for more consistent and interpretable AI-driven sentiment models.

\section{Conclusions}
\label{sec:Con}

Uncertainty and MVP in LLM emerge from a complex interplay of several factors that significantly influence the performance of sentiment analysis. These include stochastic inference mechanisms, uncertainties embedded within training datasets, architectural biases, and prompt sensitivity. A detailed review of the literature confirms that despite notable advancements, LLMs continue to exhibit instability in sentiment classification. This variability presents significant challenges in high-stakes applications that demand exceptional accuracy, reliability, and reproducibility, such as financial analytics, healthcare diagnostics, and strategic business decision-making. These concerns highlight the pressing need for robust and effective mitigation strategies.

Furthermore, the issue of sentiment analysis variability is not exclusive to contemporary deep learning approaches but is also deeply rooted in classical sentiment classification methodologies. As highlighted in \cite{wankhade2022survey}, long-standing challenges such as ambiguity, sarcasm detection, and domain-specific nuances have historically hindered sentiment classification reliability. Thus, an effective solution to sentiment variability should integrate insights from both traditional lexicon-based approaches and modern deep learning methodologies, leveraging the complementary advantages each offers.

Addressing MVP requires a multidimensional approach that incorporates advanced uncertainty quantification frameworks, structured calibration techniques, and interpretability enhancement strategies. This study underscores the need to systematically address uncertainty and variability to develop stable, interpretable, and trustworthy AI-driven sentiment analysis systems. Integration of uncertainty-aware learning methods, ensemble-based consensus strategies, domain-adaptive fine-tuning techniques, and robust explainability mechanisms is vital to mitigating model variability. Collectively, these approaches promise to enhance the reliability and consistency of sentiment classification, promoting greater acceptance and practical deployment in real-world decision-making contexts.

In summary, this study highlights how MVP, which is based on factors such as prompt design, temperature, alignment procedures, and compression, directly affects the consistency and trustworthiness of the LLM output. These challenges are particularly relevant in sentiment-driven applications, where even minor instability can misguide decision-making. Our analysis underscores the urgency of integrating stability-aware design, structured prompt strategies, and uncertainty quantification into the LLM development cycle. Looking ahead, addressing model variability is not just a research priority, but a practical requirement for responsible AI deployment, especially as LLMs are increasingly used in regulated sectors such as healthcare, finance, and public services. Establishing reproducibility standards, interpretability audits, and calibration protocols will be essential to ensure compliance with emerging governance frameworks and to maintain public trust in LLM technologies.

\textit{Novel Contributions.} This review offers several contributions to the understanding of model variability in LLMs: (1) a 12-factor taxonomy of causes contributing to output inconsistency, (2) a detailed analysis of temperature as a variability amplifier in inference, (3) dual case studies demonstrating MVP in real-world LLMs (GPT-4o and Mixtral 8x22B), and (4) mitigation strategies aligned with explainability and trust frameworks. Together, these provide both theoretical grounding and actionable practices for developing more reliable LLM pipelines.

Future work should prioritize formal benchmarks for output consistency under varied temperature settings, as well as develop general-purpose calibration frameworks for post-training stabilization. As open-source LLMs like DeepSeek, LLaMA, Mistral, or Falcon continue to proliferate, reproducibility standards and compression-aware evaluation tools will be essential in ensuring safe and reliable deployments across industry and academia.

\subsubsection*{Acknowledgements} This research results from the Strategic Project IAFER-Cib (C074/23), as a result of the collaboration agreement signed between the National Institute of Cybersecurity (INCIBE) and the University of Granada. This initiative is carried out within the framework of the Recovery, Transformation and Resilience Plan funds, financed by the European Union (Next Generation).

\bibliographystyle{apacite}

\bibliography{sample}

\end{document}